%% file: main.tex
\documentclass[twoside]{article}
\usepackage{ecj,palatino,epsfig,latexsym,natbib}
\usepackage{graphicx}
\usepackage{amsmath}
\usepackage{hyperref}

\begin{document}

\ecjHeader{x}{x}{xxx-xxx}{201X}{Diagnosing selection schemes}{Hernandez, Lalejini, and Ofria}
\title{\bf Diagnostic metrics for characterizing selection schemes}  

\author{\name{\bf Jose Guadalupe Hernandez} \hfill \addr{Jose.Hernandez8@cshs.org}\\ 
        \addr{Department of Computational Biomedicine, Cedars-Sinai Medical Center, Los Angeles, CA, USA}
\AND
    \name{\bf Alexander Lalejini} \hfill \addr{lalejina@gvsu.edu}\\      
   \addr{School of Computing, Grand Valley State University, Allendale, MI, USA}
\AND
   \name{\bf Charles Ofria} \hfill 
    \addr{ofria@msu.edu}\\
    \addr{Department of Computer Science and Engineering, Michigan State University, East Lansing, MI, USA}
}

\maketitle

\begin{abstract}

Benchmark suites are crucial for assessing the performance of evolutionary algorithms, but the constituent problems are often too complex to provide clear intuition about an algorithm's strengths and weaknesses.
To address this gap, we introduce DOSSIER (``Diagnostic Overview of Selection Schemes In Evolutionary Runs''), a diagnostic suite initially composed of eight handcrafted metrics.
These metrics are designed to empirically measure specific capacities for exploitation, exploration, and their interactions.
We consider exploitation both with and without constraints, and we divide exploration into two aspects: diversity exploration (the ability to simultaneously explore multiple pathways) and valley-crossing exploration (the ability to cross wider and wider fitness valleys).
We apply DOSSIER to six popular selection schemes: truncation, tournament, fitness sharing, lexicase, nondominated sorting, and novelty search.
Our results confirm that simple schemes (\textit{e.g.}, tournament and truncation) emphasized exploitation.
For more sophisticated schemes, however, our diagnostics revealed interesting dynamics.
Lexicase selection performed moderately well across all diagnostics that did not incorporate valley crossing, but faltered dramatically whenever valleys were present, performing worse than even random search. 
Fitness sharing was the only scheme to effectively contend with valley crossing but it struggled with the other diagnostics.
Our study highlights the utility of using diagnostics to gain nuanced insights into selection scheme characteristics, which can inform the design of new selection methods.

\end{abstract}


\input{Text/introduction}
\input{Text/diagnostics}
\input{Text/methods}
\input{Text/results}
\input{Text/conclusion}

\small

\bibliographystyle{apalike}
\bibliography{references,software}

\end{document}

%% file: Text/introduction.tex

\section{Introduction}
\label{sec:intro}


Evolutionary algorithms (EAs) are effective general-purpose techniques for solving complex problems.
Many types of EAs exist, differing in selection schemes, representations, variation operators, and other factors.
However, choosing which EA to use for a problem---let alone configuring it---remains challenging \citep{mit-ec}.
Numerous benchmarking suites are available to assess the strengths and weaknesses of each EA (\textit{e.g.}, \cite{li2013benchmark,awad2016problem,helmuth2021psb2,helmuth2015program,hansen2009real}), but do so indirectly by focusing on exemplar problems.
Here, we introduce the diagnostic suite DOSSIER (``Diagnostic Overview of Selection Schemes In Evolutionary Runs'') to measure a selection scheme's performance on different search space characteristics, starting with 
(1) the need to exploit narrow gradients, (2) the need to simultaneously explore multiple pathways, and (3) the need to cross fitness valleys.
DOSSIER initially includes eight diagnostics that each focus on each combination of these problem-solving characteristics. 
We apply these diagnostics to six commonly-used selection schemes and compare their performance.
Each diagnostic is a simple test function designed to be lightweight and intuitive, allowing it to be evaluated quickly while producing easily interpretable results.


Selection schemes determine which individuals contribute genetic material to the next generation, thus driving an EA's search strategy.
Given that problems differ in search space topology, search strategies that are effective in one search space may struggle in another.
Selection schemes vary in the criteria they use to select parents (\textit{e.g.}, performance, genetic distinctness, phenotypic rarity, \textit{etc.}) and how these criteria are used for selecting parents (\textit{e.g.}, choosing values that are best, diverse, novel, \textit{etc.}).
To choose a selection scheme that will be efficient and productive for a given problem, we must first understand the dynamics of each selection scheme on a range of search space characteristics.
For example, evolving populations might need to exploit narrow gradients, balance conflicting objectives, deal with noise, or develop building blocks to scaffold complexity.
Different selection schemes balance these capabilities in unique ways to find high-quality solutions.
Of course, these diagnostics only focus on the effectiveness of a selection scheme; a practitioner must also consider other practicalities, such as computational efficiency, implementation difficulty, or urgency of finding a ``good enough'' solution.


The common approach to understanding and comparing the problem-solving capabilities of different EAs is to use a benchmark suite with an assortment of curated challenges \citep{garden2014analysis}.
However, no standard benchmark suite exists due to the huge variety of problems that EAs are applied to \citep{jamil2013literature,hussain2017common}.
Benchmark problems can be classified into two broad categories: real-world problems and test functions.
Real-world problems are typically challenges researchers encountered ``in the wild'' and used EAs to solve \citep{helmuth2021psb2,helmuth2015program}.
These problems proved interesting, so they were chosen to provide insight into which problem domains an EA is suited for.
Test functions are well-documented mathematical functions that are fast to evaluate and usually represent idealized versions of search spaces encountered in real-world problems \citep{li2013benchmark,awad2016problem,hansen2009real}.
Additionally, test functions are often tunable, allowing researchers to expose EAs to numerous scenarios.


While benchmark suites provide useful information about particular problem domains, they have well-known limitations:
Many produce domain-specific results that benefit one algorithm over others \citep{cenikj2022SELECTOR},
or are simply not composed of particularly diverse problems \citep{garden2014analysis}.
Often, statistical analyses are inadequate \citep{lopez2021reproducibility}, and indeed,
sufficient replications for robust statistical results may be infeasible \citep{vermetten2022analyzing}.
These factors limit our ability to compare and predict how EAs will react to subtle changes in a problem.
Furthermore, most benchmark problems possess numerous integrated characteristics (\textit{e.g.}, modality, deception, separability, \textit{etc.}) that each impact problem-solving success.
The effects of specific problem characteristics cannot be disentangled without extensive experimentation and analysis, which makes it difficult to understand how an EA traverses search spaces.
More finely targeted diagnostics are required to provide deeper intuition about the strengths and weaknesses of EAs.


In this work, we apply our diagnostics to six popular selection schemes: truncation, tournament, lexicase, fitness sharing, nondominated sorting, and novelty search.
We start by assessing each selection scheme on how well it can optimize simple problems that involve unconstrained exploitation.  
We then systematically investigate three core characteristics of problem challenges: 
Constraints on the order in which exploitation can occur (``Constrained Exploitation''), 
the existence of optima in multiple neighborhoods that need to be considered simultaneously (``Parallel Exploration''), 
and the presence of fitness valleys that must be crossed (``Valley-Crossing Exploration'').
Table \ref{table:diagnostic-summary} summarizes how we convert these three characteristics into diagnostics.

\begin{table}[h!]
    \centering
    \begin{tabular}{|l|c|c|c|}
        \hline
        \textbf{Diagnostic} & \textbf{Constrained?} & \textbf{Parallel?} & \textbf{Valleys?} \\
        \hline
        1: Exploitation Rate (ER) & & & \\
        \hline
        2: Ordered Exploitation (OE) & X & & \\
        \hline
        3: Contradictory Objectives (CO) & & X & \\
        \hline
        4: Multi-Path Exploration (MP) & X & X & \\
        \hline
        5: Valley Crossing (VC) & & & X \\
        \hline
        6: Ordered Exploitation w/Valleys (OV) & X & & X \\
        \hline
        7: Contradictory Objectives w/Valleys (CV) & & X & X \\
        \hline
        8: Multiple Paths w/Valleys (MV) & X & X & X \\
        \hline
    \end{tabular}
    \caption{Diagnostic Characteristics}
    \label{table:diagnostic-summary}
\end{table}

Ultimately, these diagnostics identify meaningful differences across six prominent selection schemes, with strong evidence that they will be applicable more broadly.

%% file: Text/diagnostics.tex
\section{Diagnostics for Exploitation and Exploration}
\label{diagnostics-exploitation-and-exploration}


The ``no free lunch'' theorem asserts that no single optimization algorithm dominates all other algorithms across every possible problem instance \citep{No-Free-Lunch}.
Even limiting ourselves to ``real-world'' problems, an EA that excels in one problem domain often falters in another.
One key determinant of problem-solving success is how an EA balances exploitation and exploration \citep{eiben1998evolutionary}, where the ideal tradeoff varies by problem and by the local characteristics within a region of a search space.
Given sophisticated EAs are composed of multiple interacting components, it is difficult to isolate their individual impact \citep{vcrepinvsek2013exploration}.


Selection schemes shape this balance between exploitation and exploration, steering the search across the genetic space.
For example, a selection scheme may prioritize high-quality solutions (exploitation) or novel solutions (exploration).
Striking the right balance is crucial:
A scheme that is too exploitative will prematurely converge the population on the nearest optimum, while a scheme that is too exploratory will scatter the population across the search space, but fail to target optima.
As such, we focus on characterizing both the exploitation and exploration abilities of selection schemes.

\subsection{Diagnostic Design To Isolate Selection Schemes}
\label{diagnostic-design}


Each diagnostic uses a handcrafted search space with distinct calculated features, topology, and fitness distribution.
The problem characteristics of interest in this work include modality, deception, epistasis (interaction among genes), and dimensionality, all of which pose unique challenges \citep{weise2012evolutionary,malan2013survey,sun2014selection}.
We handcraft each diagnostic's search space from a simple common setup, avoiding complex search space topologies in favor of intuitive and interpretable designs that are intended to challenge selection schemes with targeted combinations of problem characteristics.
This design allows for differences in performance to be attributed to specific characteristics of a diagnostic. 


The design details of an EA must be considered and controlled (\textit{e.g.}, representation, variation operators, and offspring construction) to isolate a selection scheme.
All diagnostics in this work take as input a genotype representation (the ``candidate solution'') consisting of a sequence of floating-point values in the range $[0.0,100.0]$.
This constrained representation creates a well-defined search space that can be rigorously analyzed and intuitively understood.
The difficulty of each diagnostic can be adjusted by changing the range and number of values (the ``dimensionality'') in a genotype.
Here, we use $100$ as the default dimensionality.


Each diagnostic specifies a translation function of a genotype into an evaluated numerical vector (the ``phenotype'') of the same dimensionality.
We refer to the value of each position in a genotype as a ``gene'' and the value of each position in a phenotype as a ``trait''.
Each trait is maximized at the upper bound ($100.0$).
Selection schemes can either operate on traits independently where each is treated as a distinct objective, or sum all traits into a single fitness value.


Selected parents reproduce asexually, where offspring may receive point mutations to individual genes following a normal distribution ($\mathcal{N}(0.0, 1.0)$).
If a mutation would send a gene out of range, we rebound it back in (\textit{i.e.}, a mutation to $-0.7$ becomes $0.7$; a mutation to $100.7$ becomes $99.3$).
For this initial study, we intentionally avoided large-effect genetic changes such as crossover to isolate each selection scheme's ability to iteratively traverse a search space. 
In future work, we will investigate the effect of crossover on how selection schemes steer populations through search spaces.

\subsection{Diagnostics for Exploitation}


Exploitation can be described as the process of guiding a population toward an optimum within a local neighborhood of a search space \citep{beyer1998explorative}.
We consider two scenarios of exploitation: (1) unconstrained exploitation where beneficial mutations are independent of each other, and (2) constrained exploitation where some mutations will become beneficial only after other beneficial mutations occur.
Purely exploitive EAs favor high-performing solutions; they are best for search spaces with smooth gradients that lead to optima, but ineffective for rugged, deceptive, or multi-modal search spaces.
While problems rarely have gradients that lead to a global optimum from anywhere in the search space, exploitation is still critical for finding local optima.

\subsubsection{Diagnostic 1: Exploitation Rate}


In search spaces with a single, smooth gradient, exploitation alone suffices to find the global optimum.
Rapid exploitation can be critical when evaluations require substantial resources (\textit{e.g.,} compute time, memory, hardware, \textit{etc.}) and improvements to existing solutions need to be found using as few evaluations as possible.
We constructed the exploitation rate as a baseline diagnostic, intended to measure a selection scheme's capacity for exploitation through a search space that is unimodal, nondeceptive, and has independent objectives.

\begin{figure}[h!]
\centering
\includegraphics[width=\textwidth]{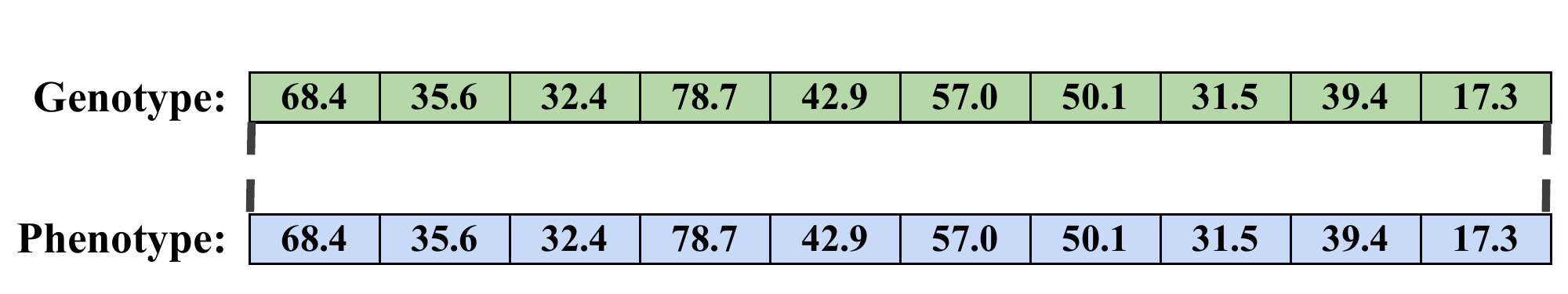}
\caption{
\textbf{An example evaluation with the exploitation rate diagnostic.} 
 A genotype of dimensionality 10 is evaluated. 
 All genes are directly copied from the genotype into the corresponding trait in the phenotype.
}
\label{fig:exploitation-eval}
\end{figure}

In this diagnostic, a genotype is directly translated to the phenotype (Figure \ref{fig:exploitation-eval}).
Because there are no interactions among genes when constructing a phenotype, this diagnostic's search space comprises multiple smooth, non-deceptive gradients (one for each trait) that can be optimized in parallel.
While this search space may be trivial to solve, it does isolate a key problem-solving characteristic: hill-climbing.
In fact, some selection schemes have known asymptotic approximations regarding exploitation \citep{goldberg1991comparative,Blickle1995ACO}, but finding this approximation is difficult for more sophisticated schemes.
Ultimately, this diagnostic allows us to compare how different schemes exploit a smooth fitness gradient.

\subsubsection{Diagnostic 2: Ordered Exploitation}


Many problems have constraints on exploitation, such as requiring one sub-problem to be solved (or advanced) before the next can be tackled. 
For example, to construct a multistory building, lower floors must be framed before progressing to higher floors. 
Framing one floor allows work to start on the next floor while the current floor is finished (\textit{e.g}., adding insulation, interior walls, \textit{etc}).  
We created the ``ordered exploitation'' diagnostic to capture a selection scheme's capacity for this type of constrained exploitation. 

\begin{figure}[h!]
\centering
\includegraphics[width=\textwidth]{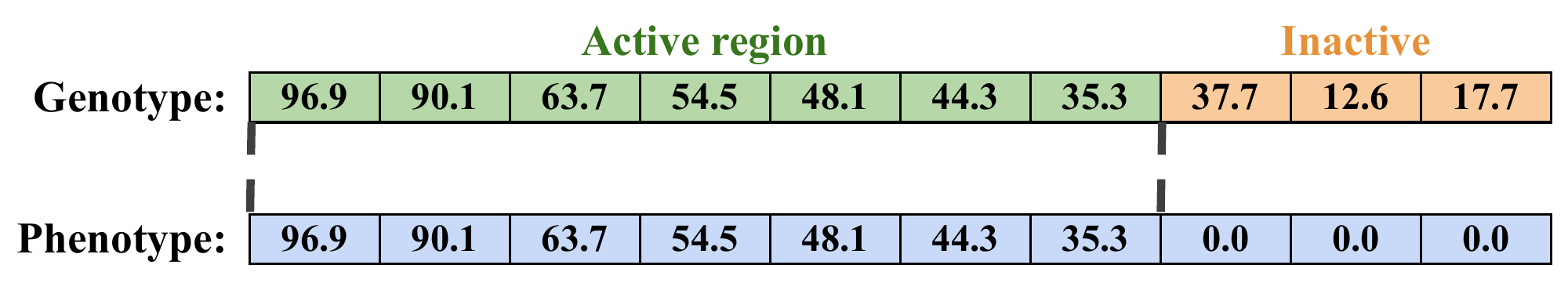}
\caption{
\textbf{An example evaluation with the ordered exploitation diagnostic.}
 A genotype of dimensionality 10 is evaluated. 
The first gene starts the active region.
It and the next six genes are all in a non-increasing sequence $(96.9$, $90.1$, $63.7$, $54.5$, $48.1$, $44.3,$ and $35.3)$, and thus define the active region. 
The next gene ($37.7$) is greater than its predecessor, and thus signifies the end of the active region. 
All genes in the active region are directly copied into the corresponding trait in the phenotype; all inactive genes are expressed as $0.0$.
}
\label{fig:ordered-exploitation-eval}
\end{figure}


In this diagnostic, genes are evaluated in order, starting from the beginning of a genotype.
The first gene is marked as ``active'', and each gene thereafter that is not greater than its predecessor is also marked as active.
The following gene that exceeds the value of its predecessor, and all subsequent genes, are marked as ``inactive''.  
We refer to the set of consecutive active genes as the ``active region''. 
All active genes are then directly interpreted as traits in the phenotype, and all inactive genes are interpreted as zero-valued traits in the phenotype (Figure \ref{fig:ordered-exploitation-eval}).


Intuitively, this diagnostic requires selection schemes to push populations through a single, narrow gradient that leads to a global optimum.
Increasing the dimensionality increases the length of the gradient, which allows us to expose schemes to more extreme, yet similar, scenarios.
This diagnostic extends the exploitation rate diagnostic by requiring that genes be optimized in order, and sufficient progress must be made on previous genes before subsequent genes can be optimized.

\subsection{Diagnostics for Parallel Exploration}


Exploration is the process of seeking new, promising neighborhoods in a search space to avert premature convergence.
For search spaces with multiple independent gradients (pathways) leading to distinct optima, a robust exploration strategy can be imperative.
Some EAs increase exploration by increasing randomness, but such an approach lacks direction.
A random walk or completely random solution will rarely gain traction in the population.
More advanced techniques maintain diverse populations by prioritizing novelty, distinctness, or Pareto optimality.
We measure a selection scheme's ability for parallel exploration both with and without constrained exploitation.
Specifically, we consider the capacity to simultaneously maintain a set of multiple equivalent optima (contradictory objectives), and the ability to simultaneously exploit multiple similar pathways to identify which one ultimately leads to a global optimum (multi-path exploration).

\subsubsection{Diagnostic 3: Contradictory Objectives}


The ``contradictory objectives'' diagnostic measures \textit{how many} global optima a selection scheme can reach and maintain in a population, providing insight into a scheme's ability to exhibit meaningful diversity.
Diversity maintenance is important for optimization problems with multiple opposing objectives, as no single optimum exists for problems with this characteristic.
Additionally, generating and maintaining a population with meaningful diversity can increase the chance of finding high-quality solutions by simultaneously exploring many distinct pathways through the search space \citep{Blickle1995ACO,vcrepinvsek2013exploration,squillero2016divergence,sudholt2020benefits}.

\begin{figure}[h]
\centering
\includegraphics[width=\textwidth]{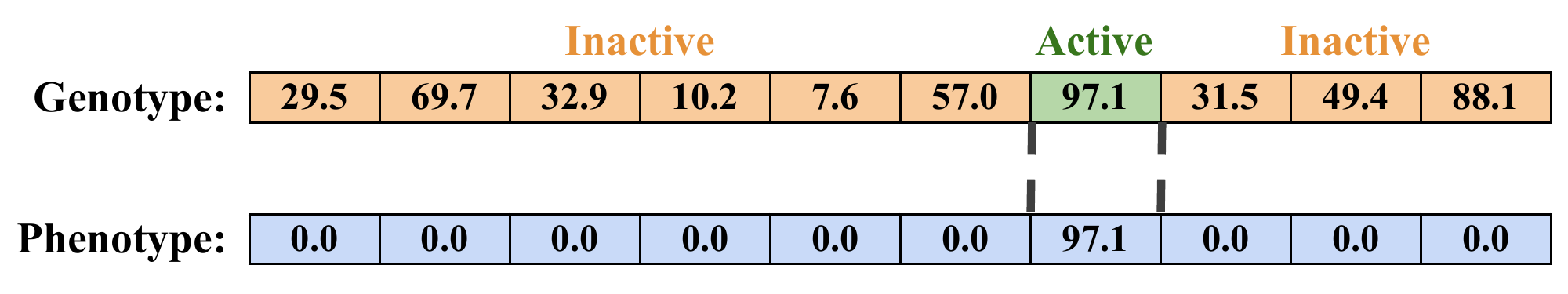}
\caption{
\textbf{An example evaluation with the contradictory objectives diagnostic.} 
A genotype of dimensionality 10 is evaluated. 
The highest gene in the genotype is identified ($97.1$) and is the only one marked as active. 
The active gene is copied into the corresponding trait in the phenotype, while
all other genes are marked as inactive and expressed as $0.0$.
}
\label{fig:contraditory-objectives-eval}
\end{figure}

Unlike the previous diagnostics, which track fitness over time, ``contradictory objectives'' is measured as a count of distinct optima that have been reached across the evolving population.
To evaluate a single genotype, we identify the gene with the greatest value (ties go to the gene closer to the start of the genotype), and we mark that gene as active.
All other genes are marked as inactive. 
The single active gene is directly interpreted into the associated trait in the phenotype, and all inactive genes are interpreted as zero-valued traits (Figure \ref{fig:contraditory-objectives-eval}).
This transformation generates one global optimum for each trait in the phenotype, where each optimum is associated with a single maximized trait.
Selection schemes must balance exploration with exploitation to discover the many gradients in the search space and to prevent the population from collapsing onto a single gradient. 

\subsubsection{Diagnostic 4: Multi-path Exploration}


The ideal tradeoff between exploitation and exploration varies by problem and by local regions of a search space.  
In contexts where multiple local optima exist, exploration can help populations discover distinct gradients, while exploitation helps populations reach those optima.
Identifying multiple local optima is often critical to finding the overall best-performing solutions.
Given the importance of selection schemes being able to navigate this balance, we constructed the  ``multi-path exploration''  diagnostic to examine the capacity for selection schemes to simultaneously explore multiple pathways and identify the global optimum.

\begin{figure}[h]
\centering
\includegraphics[width=\textwidth]{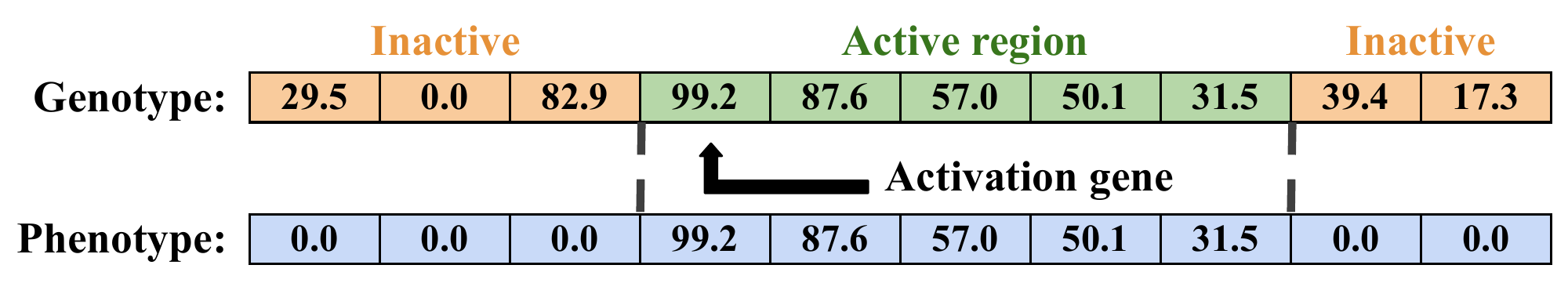}
\caption{
\textbf{An example evaluation with the multi-path exploration diagnostic.} 
A genotype of dimensionality 10 is evaluated. 
The overall highest gene is identified ($99.2$) and is set as the ``activation gene'', initiating the active region. 
The next four genes are all in a non-increasing sequence ($87.6$, $57.0$, $50.1$, and $31.5$) and are thus all included in the active region. 
The gene after the active region ($39.4$) is greater than its predecessor, thus closing the active region before it is included. 
All active genes are directly copied into the corresponding trait in the phenotype, while all other genes are marked as inactive and expressed as $0.0$.
}
\label{fig:multi-path-eval}
\end{figure}


To evaluate a genotype, we first label the highest-valued gene as the ``activation gene'' (ties go to the gene closer to the start of the genotype), and mark subsequent genes as part of the ``active region'', stopping immediately before the first gene that is greater than its predecessor.
All genes outside of the non-increasing series in the active region are marked inactive.  
Genes in the active region are directly interpreted as traits in the phenotype, while inactive genes are interpreted as zero-valued traits in the phenotype (Figure \ref{fig:multi-path-eval}).


Intuitively, this diagnostic's search space consists of multiple gradients (pathways), one for each possible activation gene.
These pathways differ in their available path length and thus peak height, but all are identical in slope.
Initially, simple beneficial mutations will exist on each path to either increase the value at the activation gene, or raise any other gene in the active region to be closer to its predecessor.
When the last gene in the active region is mutated to be greater than the subsequent gene in the genome, that next gene will join the active region, increasing the number of genes contributing to trait values.
It is much less likely, however, for an active region to extend to the left.
The activation gene is already under high pressure to be maximized, while the gene before it is under no selective pressure, making backward shifts in the activation region very unlikely.
Pathways are initially indistinguishable when the active region is small, unless the activation gene is at the end of the genome.
The potential of any given pathway can be determined only by reaching its end.
As such, the pathway beginning at the first gene in the genotype is the only one that leads to the global optimum via hill-climbing.
This diagnostic measures how well a selection scheme can simultaneously explore multiple pathways (like the contradictory objectives diagnostic) while also pursuing narrow pathways (like the ordered exploitation diagnostic).


The multi-path exploration diagnostic has already proven to be a valuable tool for analyzing selection schemes.
In \cite{hernandez2022exploration}, we used this diagnostic to produce actionable recommendations on how to maximize the exploratory capacity of lexicase selection and several of its variants.
This diagnostic has also been used to demonstrate that measures of phylogenetic diversity can provide meaningfully different information about an evolving population than measures of phenotypic diversity \citep{hernandez2022can}.

\subsection{Diagnostics for Valley-crossing Exploration}


There are many possible mechanisms for a population to explore a search space.
Parallel exploration identifies promising regions of a search space by exploring in many directions simultaneously, but it assumes that all pathways are entirely neutral or beneficial.
Nevertheless, some search spaces may require crossing fitness valleys to find higher fitness values.
Small mutational steps and crossover with a uniform population may be insufficient to cross some valleys \citep{ragusa2022augmenting,oliveto2018escape}.
Selection schemes can direct populations across a fitness valley by strategically selecting parents that occupy lower-fitness regions of a search space. 
As such, diagnostics 5 through 8 from Table \ref{table:diagnostic-summary} are valley-crossing variants for each of the previous four diagnostics.
Specifically, we integrated successively more challenging fitness valleys within the search spaces of each diagnostic to measure a selection scheme's valley-crossing exploration in all combinations with other landscape aspects.

\begin{figure}[h]
\centering
\includegraphics[width=\textwidth]{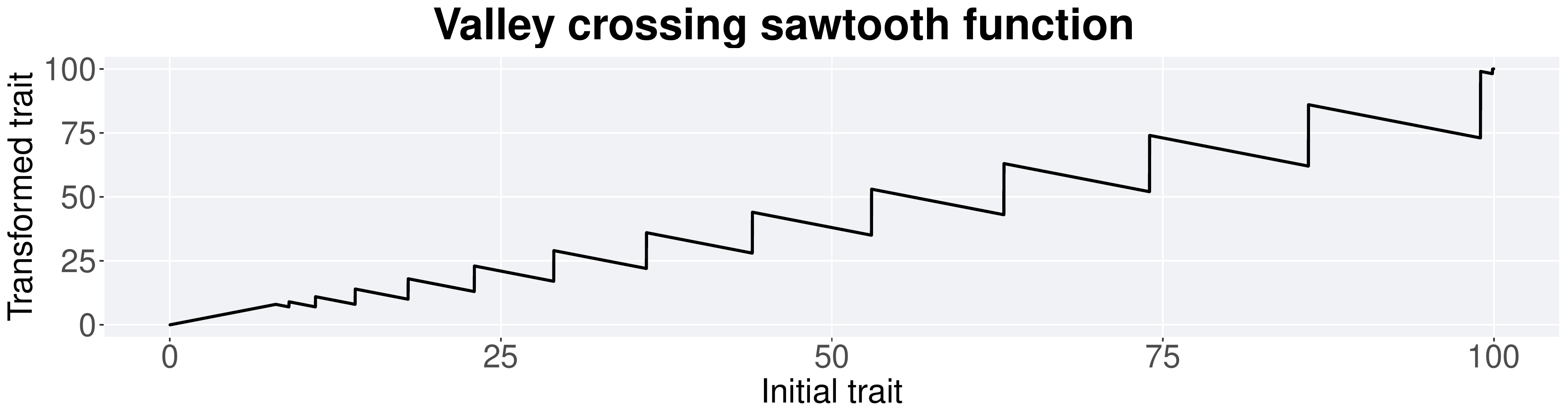}
\caption{\textbf{Valley-crossing diagnostics apply the sawtooth function to transform initial trait values.} 
The sawtooth function contains $14$ fitness peaks at gene values $8.0$ ($V_{initial}$), $9.0$, $11.0$, $14.0$, $18.0$, $23.0$, $29.0$, $36.0$, $44.0$, $53.0$, $63.0$, $74.0$, $86.0$, and $99.0$.  Each peak is followed by an ever-wider fitness valley, ending at the next peak.}
\label{fig:sawtooth-transformation}
\end{figure}


Each of the ``valley-crossing'' diagnostics applies one of the previous diagnostics, but then performs a final transformation on each would-be trait value using the sawtooth function from Figure~\ref{fig:sawtooth-transformation}.
Our sawtooth function is parameterized such that traits are allowed to initially rise linearly until they reach the initial peak value ($V_{initial}$).
After this point, valleys are introduced, starting with a width of $1$ and increasing by an additional $1$ for each subsequent valley.
Valleys descend linearly, but step back up to the untransformed value at the end of a valley, reaching a new peak.

%% file: Text/methods.tex
\section{Methods}


We conducted eight sets of experiments, each assessing how different selection schemes react to 
our diagnostics.
We compared seven popular selection schemes (truncation, tournament, genotypic fitness sharing, phenotypic fitness sharing, lexicase, nondominated sorting selection, and novelty search), and a random selection control.
In all, we had eight diagnostics each run with either selection scheme, for a total of 64 treatments.
Within each treatment, we performed $50$ replicates; for each replicate, we evolved populations of $512$ individuals for $50,000$ generations.


We initialized each population with low-fitness individuals composed of random values ranging from $0.0$ to $1.0$ (far below the 100.0 cap on genes). 
For each generation, we evaluated individuals according to the given diagnostic, and we used the target selection scheme to identify $512$ parents (with replacement) for asexual reproduction into the next generation.
We applied point mutations to an offspring's genotype at a per-gene probability of $0.7\%$.
This same process was repeated for all generations.

\input{Text/selection-schemes}

\subsection{Data Tracking}
\label{sec:methods:data-tracking}


For all diagnostics, we measure the overall quality of a search by the best-performing solution found in each generation.
We report the \textit{performance} of a solution as its average trait score, which is the sum of individual traits divided by the dimensionality (100), resulting in values between $0.0$ and $100.0$.
Additionally, we record the generation a \textit{satisfactory solution} is first discovered within the population.
We define a trait to be satisfactory if it has a value greater than or equal to $99\%$ of the upper bound ($100.0$); if all traits in a phenotype are satisfactory, we designate the corresponding solution as a ``satisfactory solution''.


For diagnostics that require parallel exploration, 
we also track both \textit{population-level satisfactory trait coverage}, which is the number of unique satisfactory traits found across all solutions in a population, and \textit{population-level activation gene coverage}, which is the number of distinct activation genes found across all solutions in a population, regardless of whether a trait is satisfactory.
Especially for the contradictory objectives diagnostic, activation gene coverage measures a scheme's capacity to produce and maintain a diverse set of phenotypes within a population,
while satisfactory trait coverage measures a scheme's ability to simultaneously exploit those mutually-exclusive traits.


The valley-crossing diagnostics assess a selection scheme's ability to cross fitness valleys integrated within other diagnostic search spaces.
We track the same metrics for those diagnostics paired with the valley-crossing diagnostic.
Additionally, we record the largest valley reached within an individual trait for the best-performing solution.

\subsection{Statistical Analysis}

We performed a Kruskal–Wallis test to determine if significant differences among selection schemes occurred.
For comparisons where the Kruskal–Wallis test was significant (significance level of 0.05), we performed a post-hoc Wilcoxon rank-sum test between relevant schemes with a Bonferroni correction for multiple comparisons where appropriate; comparisons that focus on performance used the sum of individual traits. 
Because novelty search uses an archive to track novel behaviors, we also consider archive solutions when generating results.

\subsection{Software Availability}
\label{sec:data-and-software-availability}

Our supplemental material~\citep{supplemental_material} is hosted on \href{https://github.com/jgh9094/ECJ-2023-Suite-Of-Diagnostic-Metrics-For-Characterizing-Selection-Schemes}{GitHub} and contains the software, data analyses, and documentation for this work.
Our experiments are implemented in C++ using the Empirical library~\citep{charles_ofria_2020_empirical}, and we used a combination of Python and R version 4 \citep{r_lang} for data processing and analysis.
The following R packages are used for data wrangling, statistical analysis, graphing, and visualization: 
ggplot2 \citep{R-ggplot2}, 
cowplot \citep{R-cowplot},
pupillometryR \citep{R-pupillometryR},
and dplyr \citep{R-dplyr}.
We used R markdown \citep{R-markdown} and bookdown \citep{R-bookdown} to generate web-enabled supplemental material.
All data is available on the Open Science Framework at \url{https://osf.io/qz3g7/}.

%% file: Text/selection-schemes.tex
\subsection{Selection Schemes}
\label{sec:selection-schemes}

In this work, we diagnose the following six selection schemes because of their popularity and demonstrated effectiveness on different problems.
Schemes that require a single fitness value use \textit{total fitness}, which is the sum of all of an individual's traits.

In many of the cases below, we needed to choose parameters for each of these selection schemes and tried to base our decisions on typical choices in the literature.
While a more thorough analysis of parameter values could be fruitful, we believe the results presented here provide reasonable baselines.
We included additional replicates that explore different parameter configurations in our supplemental material \citep{supplemental_material}.

\subsubsection{Truncation Selection}
\label{sec:truncation}


Truncation selection uses the top performing (``elite'') solutions in a population as parents to create the next generation.
This selection scheme is the simplest and most direct way to identify high-quality parents; it is the basis for most animal husbandry \citep{Crow396} and is used widely within evolutionary computation \citep{beyer2002evolution,Luke2013Metaheuristics}.

Selection is performed by sorting the population by total fitness (with ties settled randomly) and then truncating it, leaving only the top $tr$ performers to be used as parents.
Selected parents produce equal numbers of offspring until the next generation contains the correct number of candidate solutions. 
Here, we use $tr=8$, meaning that after truncation, each parent creates $64$ offspring.


The level of truncation, $tr$, dictates the strength of selection \citep{goldberg1991comparative,back1996extending}.
At one extreme, if $tr=1$, selection pressure is maximal and the single highest fitness solution gives rise to the entire next generation.
At the other extreme, if $tr$ is the population size, each individual produces one offspring and is on a random walk.
Our choice of $tr=8$ is a mid-range value commonly used in the existing literature, biased toward exploitation.
 
\subsubsection{Tournament Selection}
\label{sec:Tournament}


Tournament selection \citep{brindle1980genetic} is one of the most commonly used selection schemes \citep{goldberg1991comparative,Luke2013Metaheuristics}. 
Each parent is chosen as the individual with the highest total fitness from a randomly selected group (tournament) of $ts$ solutions.
Each tournament identifies a single parent, thus 512 (population size) tournaments are held given our setup.


The tournament size, $ts$, determines the strength of selection \citep{goldberg1991comparative,back1996extending}.  
As $ts$ approaches 1, tournament selection behaves more like random selection, allowing low-fitness solutions to become parents.
As the $ts$ approaches the population size, only the most fit individuals are able to win tournaments.
Here we set the tournament size to an intermediate size of $ts=8$, a common value used in the literature.

\subsubsection{Fitness Sharing}
\label{sec:fitness-sharing}


Fitness sharing incorporates an explicit mechanism for maintaining a diverse population to mitigate premature convergence \citep{goldberg1987genetic}.
At the start of the selection step, all candidate solutions are assigned a similarity score, which is a solution's likeness to the rest of the population.
The similarity metric can be either genotypic (\textit{e.g.}, the hamming distance between genomes) or phenotypic (\textit{e.g.}, the number of traits two individuals have in common).
Fitness sharing modifies each solution's total fitness, decreasing it as a function of its similarity to the rest of the population; solutions occupying crowded regions of the search space or solution space have their fitness reduced more than those in less crowded regions.


Consider candidate solution $x$ with $f_x$ representing the sum of all of its traits after being evaluated on a diagnostic.
The shared fitness $f'_{x}$ of solution $x$ is given by
\[ 
    f'_{x} = \frac{f_x}{m_x}
\]
where $m_x$ quantifies solution $x$'s fitness reduction due to its similarity to the rest of the population. 
Here, we use two variations of fitness sharing, one that uses genotypic similarity and one that uses phenotypic similarity, which we refer to as genotypic fitness sharing and phenotypic fitness sharing.
For both variations, we calculate $m_{x}$ as 
\[ 
    m_{x} = \sum_{y \in P} S(d_{xy}) 
\]
where $P$ is the population, $S()$ is the sharing function, and $d_{xy}$ is the Euclidean distance between the genotypes or phenotypes for solutions $x$ and $y$.
The sharing function $S()$ uses the distance between two solutions to set the associated fitness penalty, if any:
\[ S(d) = 
    \begin{cases} 
        1 - (\frac{d}{\sigma})^{\alpha}, & \text{if } d < \sigma \\
        0, & \text{otherwise}
   \end{cases}
\]

Two variables are required to configure the sharing function: $\alpha$ and $\sigma$.
The parameter $\alpha$ regulates the shape of the sharing function and $\sigma$ determines the threshold of dissimilarity beyond which no penalty should exist.
We use $\alpha = 1.0$ and $\sigma = 0.3$ for all replicates in this work.
We selected this $\alpha$ value due to it being commonly used \citep{goldberg1987genetic,sareni1998fitness} and we empirically identified a generally effective $\sigma$ value~\citep{supplemental_material}.
Once all solutions have their shared fitness assigned, the stochastic remainder selection with replacement described in \cite{haq2019novel} is used to identify parents, as it is recommended to pair fitness sharing with stochastic remainder selection \citep{goldberg1987genetic,sareni1998fitness}.

\subsubsection{Lexicase Selection}
\label{sec:lexicase}


Lexicase selection is a technique designed for genetic programming problems where solutions must perform well across multiple test cases \citep{helmuth2014solving,orzechowski2018where,helmuth2020benchmarking}.
The previously described selection schemes focus on maximizing a fitness value (truncation, tournament selection) or promoting uniqueness (fitness sharing).
By contrast, lexicase selection selects for individuals that specialize on different \textit{combinations} of test cases by iterating through shuffled sets of test cases, resulting in high levels of stable diversity \citep{helmuth2016effects,dolson2018ecological,helmuth2020importance}.


In lexicase selection, all candidate solutions are evaluated on a set of test cases, and their performance on each test case is recorded.
Here, we associate one test case with each trait in a solution's phenotype (resulting in 100 test cases).
We use each trait value in a solution's phenotype as a direct measure of performance for the associated test case.
To identify a parent, lexicase selection shuffles the set of test cases and iterates through each test case in sequence.
Starting from the full population, each test case is used to filter down the current set of candidate parents; only those solutions tied for best performance on a given test case are allowed to continue.
This filtering process continues until a only single solution remains or all test cases have been used.
If multiple candidates remain, a random one is selected to be the parent.


Lexicase selection has no free parameters to set.
It is designed to balance both exploitation and exploration without requiring a user to intervene by specifying what that balance should look like.

\subsubsection{Nondominated Sorting}
\label{sec:nondominated-sorting}


The nondominated sorting genetic algorithm (NSGA) \citep{srinivas1994muiltiobjective} and its descendants \citep{NSGA-II,NSGA-III} are successful evolutionary multi-objective optimization techniques. 
Evolutionary multi-objective optimization methods aim to generate a set of solutions that represent the best possible tradeoffs among multiple (possibly conflicting) objectives~\citep{coellocoello_evolutionary_2020}.
NSGA combines two procedures for selection: a ranking procedure that groups individuals into nondominated fronts and fitness sharing for diversity maintenance within each front.


We use the set of phenotypes produced by our diagnostics to identify whether or not a solution is dominated in a given population.
Phenotypes are compared on a trait-by-trait basis.
Given two phenotypes $x$ and $y$, we say that $x$ \textit{dominates} $y$ if all of $x$'s traits are greater than or equal to the corresponding trait in $y$ and at least one of $x$'s trait is strictly greater than the trait in $y$.


The first nondominated front is created by collecting all the solutions not dominated by any other solution in the population.
Each subsequent front is constructed by finding the set of nondominated solutions from the solutions not yet part of any front.
As each front is constructed, all solutions within that front are initially assigned the same fitness value.
Fitness sharing is then applied to the solutions within each front; the same procedure in Section \ref{sec:fitness-sharing} is used to calculate shared fitness with phenotypic similarity.
The initial fitness value for each front is selected such that it is less than all shared fitness values from previous fronts, and the procedure is repeated.
Once all solutions have their shared fitness finalized, stochastic remainder selection (see Section \ref{sec:fitness-sharing}) is used to identify parents.

\subsubsection{Novelty Search}
\label{sec:novelty-search}


Novelty search mitigates complications associated with objective functions (\textit{e.g.,} deception and local optima) by abandoning traditional fitness-based objectives \citep{lehman2008exploiting}, instead selecting solutions based on behavioral distinctiveness.
Novelty search uses novelty scores to preferentially select solutions with behaviors distinct from those previously observed, encouraging ``productive'' exploration without any explicit focus on optimality.


We use the phenotype returned by our diagnostics to represent the set of behaviors used for measuring novelty.
Consider a phenotype $x$ outputed from some diagnostic.
The novelty score of $x$ is given by
\[ 
    \rho(x) =  \frac{1}{k} \sum_{i=0}^{k} dist(x, u_i)
\]
where $\rho(x)$ is the novelty score of $x$, $dist()$ is the distance function, and $u_i$ is the $i$-th nearest neighbor of $x$ with respect to phenotypes.
All calculations of nearest-neighbor phenotypes include both the current population and an archive of all novel phenotypes previously found. 
Here, we used the Euclidean distance between two phenotypes as the distance metric.
We also set $k=15$, as recommended in \cite{lehman2008exploiting}.
Once the novelty scores are calculated for all candidate solutions, we used tournament selection with size two to identify parents with high novelty scores (as in \cite{lehman2010efficiently} and \cite{jundt2019comparing}).


Since novelty search is focused on finding phenotypes that were never previously encountered, maintaining an unbounded archive is important.
We use a threshold $pmin$ to determine whether a phenotype is sufficiently novel to be tracked by the archive.
Here, $pmin$ is set to $10.0$.
Approximately one phenotype is randomly saved to the archive every 200 generations.
If more than $4$ phenotypes enter the archive by being more novel than $pmin$ in one generation, $pmin$ is increased by $25\%$.
If no new phenotypes are added to the archive for $500$ generations, $pmin$ is decreased by $5\%$.
This configuration closely follows the novelty search used in \cite{lehman2008exploiting}.

%% file: Text/results.tex
\section{Results and Discussion}

\subsection{Exploitation Diagnostics}

\begin{figure}[h]
\centering
\includegraphics[width=\textwidth]{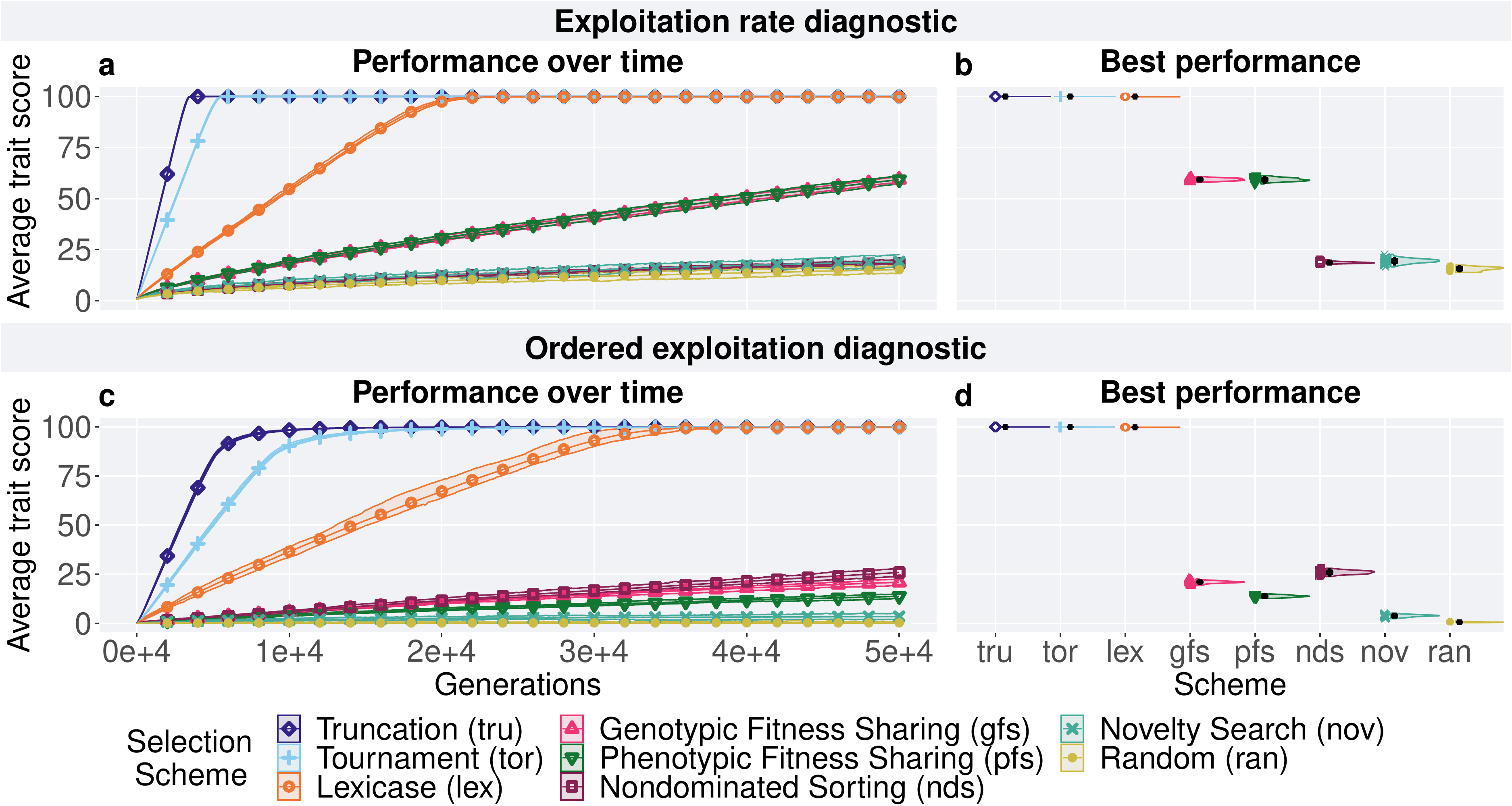}
\caption{
\textbf{Results for selection schemes evaluated on the exploitation rate and ordered exploitation diagnostics.} 
Best performance in the population (a, c) over time and (b, d) earned throughout $50,000$ generations. 
For panels (a) and (c), we plot the average, best, and worst performances (with shading in between) across the 50 replicates every 2000 generations.
}
\label{fig:res:exploitation-focused}
\end{figure}

Using both the exploitation rate and ordered exploitation diagnostics, we compare the relative ability to exploit a fitness gradient among selection schemes.
We found that all schemes improve performance over time and outperform the random control when comparing the best performance found throughout an evolutionary run on both diagnostics (Figure \ref{fig:res:exploitation-focused}b and \ref{fig:res:exploitation-focused}d; Wilcoxon rank-sum test: $p<10^{-4}$).
Indeed, the schemes that prioritize exploitation succeed in both diagnostics.


Only truncation, tournament, and lexicase selection reached satisfactory solutions for both diagnostics by $50,000$ generations, doing so in all replicates.
For the configurations used here, truncation selection found satisfactory solutions in the fewest generations, while tournament selection found satisfactory solutions in fewer generations than lexicase selection (Wilcoxon rank-sum test: $p<10^{-4}$).
This result is consistent with the literature, as truncation selection exhibits stronger selective pressure than tournament selection \citep{goldberg1991comparative,back1996extending}.
Evidently, both truncation and tournament selection reach satisfactory solutions faster than lexicase selection due to maximizing an aggregate score, whereas lexicase selection pressures the population to be best at multiple test case combinations.
Lexicase selection's diversity maintenance does not provide any explicit advantage for solutions with larger aggregate scores, slowing down the rate of exploitation on both diagnostics.


Both fitness sharing treatments produce identical results for the exploitation rate diagnostic; no statistical difference is found between their best performances (Figure \ref{fig:res:exploitation-focused}b; Wilcoxon rank-sum test: $p>0.05$).
Given that the exploitation rate diagnostic directly copies a genotype into a phenotype, there is no procedural difference between both fitness sharing treatments.
As populations can maximize traits in any order, fitness sharing penalizes individuals maximizing similar traits, lowering the chances of solutions with larger aggregate scores from being selected.
Nonetheless, both fitness sharing treatments reached better-performing solutions than nondominated sorting and novelty search (Wilcoxon rank-sum test: $p<10^{-4}$).


Interestingly, genotypic fitness sharing found better-performing solutions than phenotypic fitness sharing on the ordered exploitation diagnostic (Figure \ref{fig:res:exploitation-focused}d; Wilcoxon rank-sum test: $p<10^{-4}$).
We suspect this dynamic occurred because the tail ends of the genotypes were non-active regions; they could drift apart when genotypes were being analyzed for fitness sharing, but were locked at zero when phenotypes were compared.
As such, genotypic fitness sharing's ability to minimize similarity allows it to outperform phenotypic fitness sharing.
For this diagnostic, both fitness sharing treatments reached better-performing solutions than novelty search (Wilcoxon rank-sum test: $p<10^{-4}$).

Nondominated sorting performs poorly on both diagnostics, as it emphasizes exploration over exploitation.
In fact, NSGA-II extends the nondominated sorting in NSGA by incorporating elitism to address this issue \citep{NSGA-II}.
However, nondominated sorting found better-performing solutions than both fitness sharing treatments and novelty search on the ordered exploitation diagnostic (Figure \ref{fig:res:exploitation-focused}d; Wilcoxon rank-sum test: $p<10^{-4}$).
We suspect nondominated sorting's performance can be explained by its focus on finding Pareto-optimal solutions.
Nondominated sorting pressures solutions to be nondominated, as solutions in early nondominated fronts have higher fitness.
This pressure favors unlocking new active genes and selecting for increases in active gene streaks.
Long streaks of active genes are found in high-performing solutions, as better-performing solutions can be reached, which helps explain its performance. 
Indeed, we found that nondominated sorting evolved longer streaks than both fitness sharing configurations and novelty search when comparing the longest streak from the best-performing solutions (\cite{supplemental_material}; Wilcoxon rank-sum test: $p<10^{-2}$).


Finally, as expected, novelty search performs poorly due to explicitly ignoring exploitation.
Other variations of novelty search may perform better as they incorporate mechanisms to increase exploitation \citep{novelty_exploit}.
Interestingly, novelty search found better-performing solutions than nondominated sorting for the exploitation diagnostic (Figure \ref{fig:res:exploitation-focused}b; Wilcoxon rank-sum test: $p<10^{-1}$).
Given enough time, however, we might expect novelty search to find satisfactory solutions by exhaustively enumerating the search space \citep{doncieux_novelty_2019}.

\subsection{Parallel Exploration}


The contradictory objectives diagnostic examines a selection scheme's capacity for parallel exploration in isolation.
Specifically, it limits each solution's phenotype to specializing on a single trait, allowing us to compare the relative ability to locate and optimize conflicting objectives across selection schemes.
For each selection scheme, we compared population-level activation gene coverage (\textit{i.e.}, the number of distinct activation genes maintained in the population) and satisfactory trait coverage (\textit{i.e.}, the number of distinct satisfied traits across the whole population).

\begin{figure}[h]
\centering
\includegraphics[width=\textwidth]{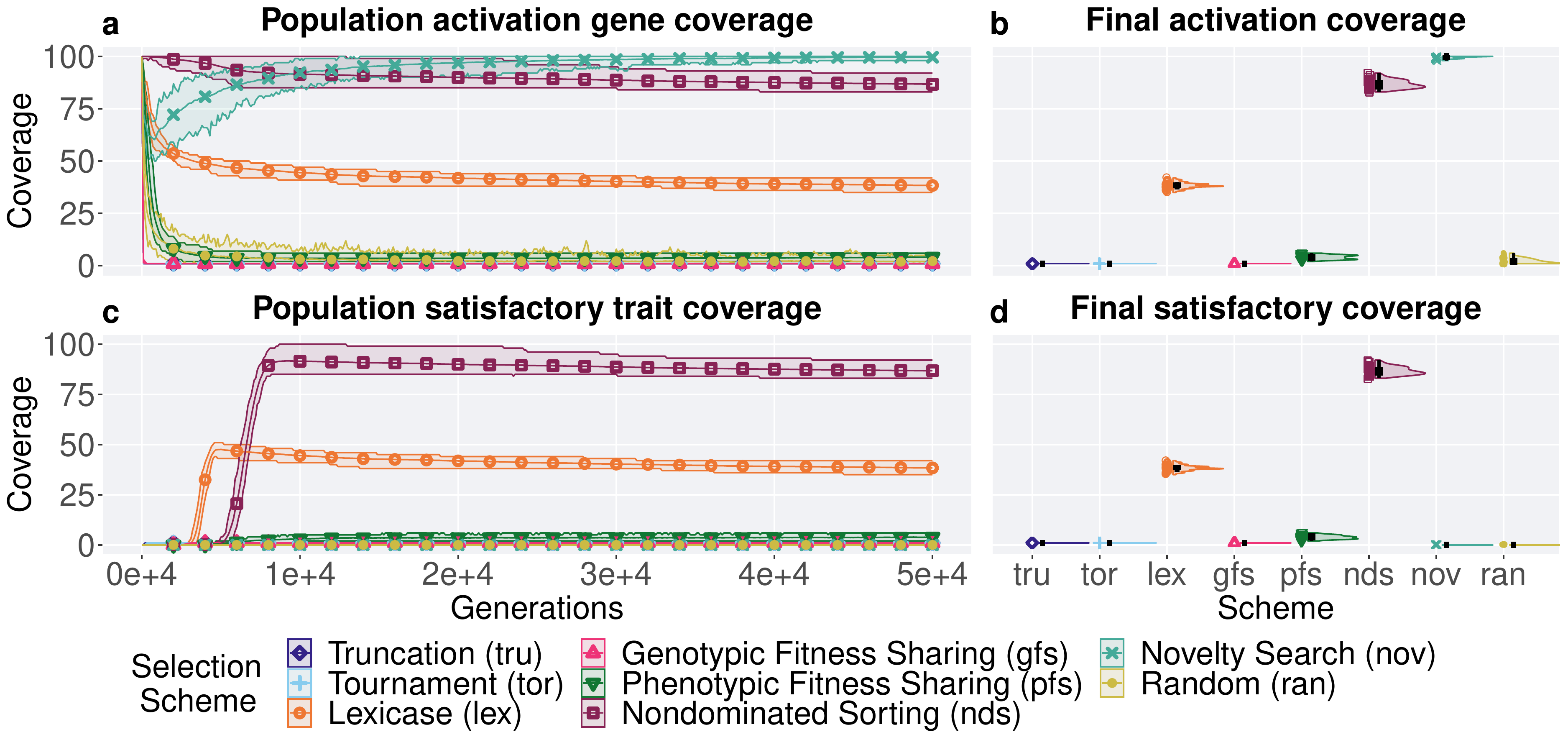}
\caption{
\textbf{Results for selection schemes evaluated on contradictory objective diagnostic.}
Population-level activation gene coverage (a) over time and (b) at $50,000$ generations. 
Population-level unique satisfactory traits (c) over time and (d) at $50,000$ generations. 
For panels (a) and (c), we plot the average, best, and worst coverage (with shading in between) across the 50 replicates every 2000 generations.
}
\label{fig:res:contradictory-objectives}
\end{figure}


Since we initialized starting populations with genotypes consisting of random values between $0.0$ and $1.0$, each initial solution has a random activation gene leading to high activation gene coverage at the start of all treatments (Figure \ref{fig:res:contradictory-objectives}a).
However, activation gene coverage rapidly decreased for all selection schemes, except nondominated sorting.
In fact, tournament selection, truncation selection, and genotypic fitness sharing consistently collapsed to a single activation gene; all other schemes, including all but one replicate of our random control, maintained multiple activation genes in each population (Figure \ref{fig:res:contradictory-objectives}b).
Over time, novelty search recovered its activation gene count, ultimately maintaining more than nondominated sorting (Wilcoxon rank-sum tests: $p<10^{-4}$).
Nondominated sorting, in turn, maintained more activation genes than lexicase selection, and lexicase selection maintained more activation genes than phenotypic fitness sharing (Wilcoxon rank-sum tests: $p<10^{-4}$).


Not all activation genes ended up being satisfied, though all selection schemes, except novelty search and the random control, did manage to satisfy at least one trait after $50,000$ generations (Figure~ \ref{fig:res:contradictory-objectives}d).
All populations evolved under truncation selection, tournament selection, and genotypic fitness sharing produced exactly one satisfactory trait in the final population.
Phenotypic fitness sharing, lexicase selection, and nondominated sorting consistently produced populations with more than one unique satisfactory trait.
For the configurations used here, nondominated sorting maintained the most satisfactory traits, while lexicase selection maintained more satisfactory traits than phenotypic fitness sharing (Wilcoxon rank-sum tests: $p<10^{-4}$).
We found that all schemes, except novelty search, attained more satisfactory traits than our random control (Wilcoxon rank-sum tests: $p<10^{-4}$).


In general, selection schemes capable of maintaining populations with diverse activation genes were also able to optimize those genes to satisfactory levels.
Note that in such cases, the diverse activation genes were typically optimized in parallel.


Novelty search did not obtain any satisfactory traits, likely due to emphasizing novel traits while not directly considering improvements to existing traits.
Clearly, pressure for novelty does not facilitate reaching a satisfactory trait within the allotted time; yet this pressure did allow novelty search to maintain higher activation gene coverage than all other selection schemes in the final population (Figure \ref{fig:res:contradictory-objectives}b; Wilcoxon rank-sum tests: $p<10^{-4}$).
We suspect this result was due to a combination of the emphasis on finding novel behaviors and the implicit diversity enhancements provided by the archive.
Because novelty search uses phenotypic similarity, the population is pressured to optimize different traits, as doing so increases novelty scores.
Early in the evolutionary search, activation gene coverage drops, yet it reaches perfect coverage by the end.
This drop in coverage occurred due to the population requiring time to diversify, as the starting novelty threshold is too high for solutions to be added to the archive.


\begin{figure}[h]
\centering
\includegraphics[width=\textwidth]{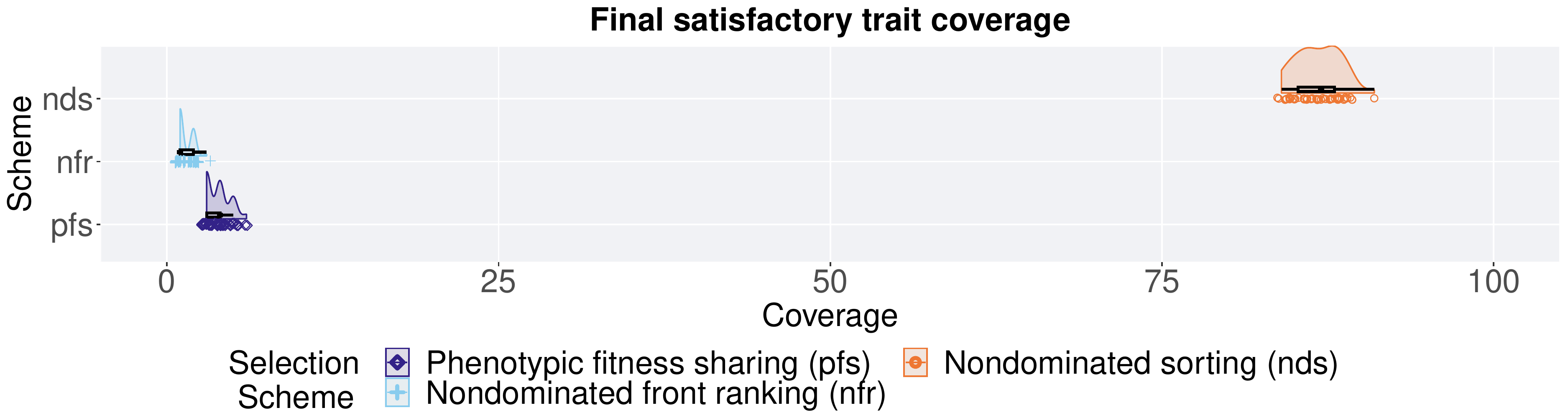}
\caption{
\textbf{Results for nondominated sorting, phenotypic fitness sharing, and nondominated front ranking evaluated on the contradictory objective diagnostic.}
We report the satisfactory trait coverage found in the final population. 
}
\label{fig:res:contradictory-objectives:nondominated}
\end{figure}

Nondominated sorting surpassed all other selection schemes at producing populations with high satisfactory trait coverage (Figure \ref{fig:res:contradictory-objectives}c and \ref{fig:res:contradictory-objectives}d; Wilcoxon rank-sum tests: $p<10^{-4}$).
This result is expected, as the contradictory objectives diagnostic generates the ideal search space for nondominated sorting, containing one equidistant Pareto-optimal solution per trait.
Nondominated sorting's performance is due to both of its diversity maintenance mechanisms: nondominated front ranking and phenotypic fitness sharing.
To illuminate the relative importance of both components, we applied nondominated front ranking (nondominated sorting with $\sigma=0.0$) and phenotypic fitness sharing to the contradictory objectives diagnostic (Figure \ref{fig:res:contradictory-objectives:nondominated}).
There is a significant drop in the satisfactory trait coverage maintained in the final population between using both nondominated front ranking and fitness sharing, as compared to using just one (Wilcoxon rank-sum tests: $p<10^{-4}$).
From this, we conclude that nondominated sorting is more than the sum of its parts; that is, both front ranking and fitness sharing were critical to its overall ability to find and maintain many optima.
Indeed, this diagnostic helps illuminate the importance of the multiple components that make up a selection scheme.


Aside from nondominated sorting, lexicase selection was the only other selection scheme to produce populations with high satisfactory trait coverage (Figure \ref{fig:res:contradictory-objectives}d).
Lexicase selection's success is consistent with previous theoretical and experimental findings that demonstrate its ability to produce populations with meaningful diversity without impeding simultaneous exploitation~\citep{helmuth2016effects,helmuth2020importance}.
Lexicase selection's emphasis on selecting specialists is valuable for performing well on this diagnostic, as a population with high satisfactory trait coverage \textit{must be} a population of specialists; given this, why does lexicase selection maintain substantially lower satisfactory trait coverage than nondominated sorting? 
Previous work has shown that lexicase selection's capacity to maintain a given specialist is related to the probability that its associated test cases appear first in the shuffles during selection~\citep{dolson2018ecological,hernandez2022exploration}. 
That is, lexicase selection is sensitive to the ratio between population size and the number of test cases. 
As such, we expect that increasing population size or decreasing diagnostic dimensionality would reduce the performance gap between nondominated sorting and lexicase selection on this diagnostic.   


Neither fitness sharing variants produced populations with high satisfactory trait coverage.
However, of these two methods of fitness sharing, phenotypic fitness sharing surpassed genotypic fitness sharing (Figure \ref{fig:res:contradictory-objectives}d; Wilcoxon rank-sum test: $p<10^{-4}$).
We suspect that this difference in outcome is driven by the information captured by each similarity metric.
Phenotypic similarity is more likely to penalize solutions that optimize the same trait, which results in greater selection pressure to optimize unique traits.  
This pressure is masked with genotypic similarity, as the similarity between solutions optimizing the same trait can be negated by inactive genes drifting. 
Thus, when comparing two solutions, genotypic fitness sharing does not focus only on the traits that those two solutions are optimizing, but this is exactly what happens with phenotypic fitness sharing.
Previous work has shown the threshold of dissimilarity and population size affect fitness sharing's ability to fill multiple niches~\citep{della2004role}; indeed, we find evidence of this, as increasing $\sigma$ leads to higher activation gene and satisfactory trait coverage (\cite{supplemental_material}; Wilcoxon rank-sum tests: $p<10^{-4}$).


Both truncation and tournament selection performed poorly on this diagnostic, as both schemes do not maintain or generate diverse populations, and exhibit strong selection pressure \citep{goldberg1991comparative,back1996extending,helmuth2016effects,hernandez2022exploration}.
Tournament selection increases the number of unique parents identified through tournaments, yet only one satisfactory trait and activation gene is maintained in the final population for all replicates.
Each selection scheme's takeover time suggests that early high-performing solutions will reduce the number of unique traits being optimized in the population.
Additionally, aggregating traits makes it impossible to differentiate what trait is being optimized.

\subsection{Balancing Constrained Exploitation and Parallel Exploration}
\label{sec:res:multi-path}


The multi-path exploration diagnostic generates a search space with multiple gradients, equal in slope but differing in length, and thus final peak fitness.
This search space allows us to compare the relative ability of selection schemes to maintain and simultaneously exploit different gradients, with the goal of fully traversing the gradient that leads to the global optimum.

\begin{figure}[h]
\centering
\includegraphics[width=\textwidth]{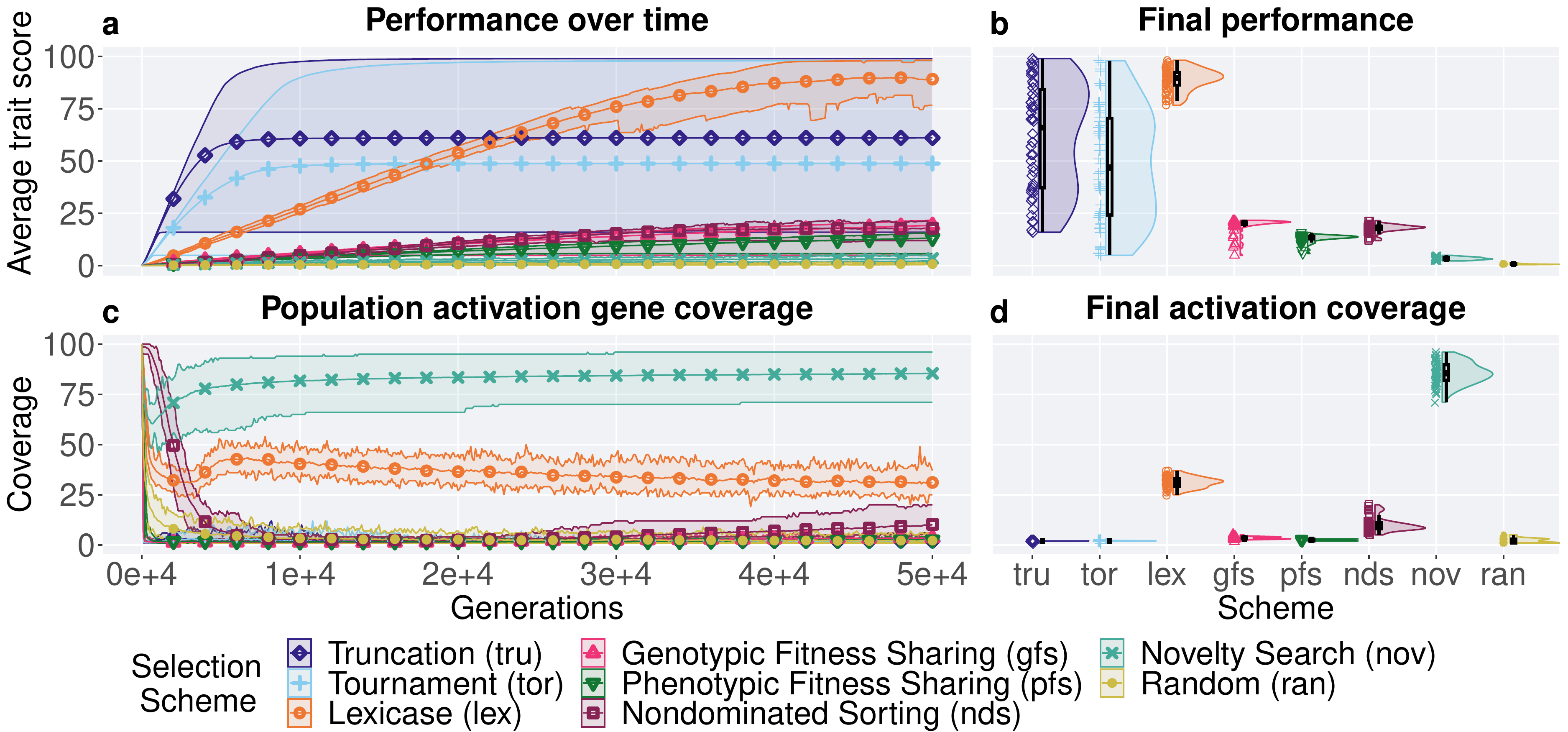}
\caption{
\textbf{Results for selection schemes evaluated on the multi-path exploration diagnostic.} 
Best performance in the population (a) over time and (b) at $50,000$ generations. 
Population-level activation gene coverage (c) over time and (d) at $50,000$ generations. 
For panels (a) and (c), we plot the average, best, and worst data (with shading in between) across the 50 replicates every 2000 generations.
}
\label{fig:res:exploration}
\end{figure}


This diagnostic proved to be challenging for all selection schemes.
While all of them improved performance over time (Figure~ \ref{fig:res:exploration}a), none were able to evolve a satisfactory solution in the allotted time.
Tournament, truncation, and lexicase selection each produced high-performing solutions and all schemes found better-performing solutions than our random control (Figure~ \ref{fig:res:exploration}b; Wilcoxon rank-sum tests: $p<10^{-4}$).
While tournament and truncation selection found some high-performing solutions, only a few replicates were able to do so.
As these selection schemes are unable to maintain parallel exploration, they should have approximately a 1\% chance of stumbling on the optimal trajectory; our results are consistent with this expectation.
Lexicase selection, by contrast, consistently reached high-performing solutions, outperforming all other schemes (Wilcoxon rank-sum tests: $p<10^{-4}$).


As with the contradictory objectives diagnostics, all initial, randomly-generated populations had high activation gene coverage that rapidly decreased (Figure \ref{fig:res:exploration}c and \ref{fig:res:exploration}d).
We found no difference between activation gene coverage in the final populations when comparing truncation selection, tournament selection, and phenotypic fitness sharing to our random control (Wilcoxon rank-sum tests: $p>0.05$); while all other selection schemes maintain more activation genes than our random control (Wilcoxon rank-sum tests: $p<10^{-4}$).
As in the previous diagnostic results, we found that novelty search, lexicase selection, and nondominated sorting maintained higher activation gene coverage than all other selection schemes in the final populations (Wilcoxon rank-sum tests: $p<10^{-4}$). 
For the configuration used here, novelty search maintained the most activation genes, while lexicase selection maintained more than nondominated sorting (Wilcoxon rank-sum tests: $p<10^{-4}$). 
All other schemes maintained low levels of activation gene coverage.


Lexicase selection was the only selection scheme to consistently reach high-performing solutions throughout an evolutionary search (Figure \ref{fig:res:exploration}a), while also maintaining high activation gene coverage (Figure \ref{fig:res:exploration}c).
These results are expected, as the previous diagnostics allow us to estimate a scheme's potential on this diagnostic.
Lexicase selection's performance on the ordered exploitation diagnostic demonstrates its ability to exploit gradients similar to those found in this diagnostic's search space, while its performance on the contradictory objectives diagnostic demonstrates its ability to maintain a diverse set of activation genes. 
Ultimately, lexicase selection is the only selection scheme to exhibit the levels of exploration and exploitation needed to reach high-performing solutions.
Lexicase selection's strong performance on the exploration diagnostic is consistent with previous work investigating the exploratory capacity of lexicase selection and its variants~ \citep{hernandez2022exploration}.   


Success on the contradictory objectives diagnostic predicts success on this diagnostic, as the ability to maintain a diversity of activation genes increases a selection scheme's chance of exploring the pathway to the global optimum.
While truncation and tournament selection are apt at exploiting gradients, neither scheme can maintain high activation gene coverage, limiting their ability to explore multiple gradients in the search space. 
Indeed, only replicates where the population (by chance) converges to a high-potential gradient produce high-fitness solutions.   


Exploitation is also crucial for success on this diagnostic, as the only way to reach the global optimum is by effectively exploiting the optimum's gradient.
While fitness sharing, nondominated sorting, and novelty search maintained multiple activation genes, they failed to fully exploit the associated gradients, which is consistent with their performances on the ordered exploitation diagnostic.

\subsection{Valley-crossing Diagnostics}

For each of the previous four diagnostic search spaces, we also examined the associated valley-crossing variation.
As expected, fitness valleys increased the difficulty of search space traversal for all diagnostics.
None of the selection schemes reached the global optimum for any of the valley-crossing diagnostics, with the sole exception of phenotypic fitness sharing on diagnostic 7: contradictory objectives with valleys.
Since the contradictory objectives diagnostics generate only a single non-zero trait, phenotypic fitness sharing focuses on that trait pushing it through valleys until it is maximized.

\begin{figure}[h]
\centering
\includegraphics[width=\textwidth]{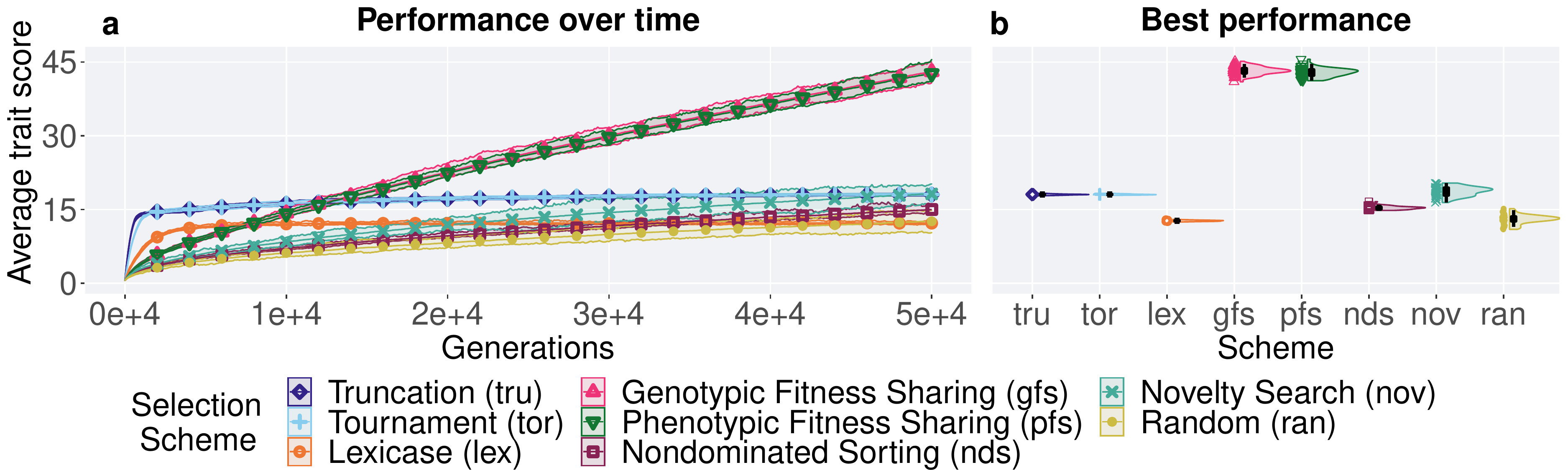}
\caption{
\textbf{Results for selection schemes evaluated on the valley-crossing diagnostic.}
Best performance in the population (a) over time and (b) throughout $50,000$ generations. 
For panel (a), we plot the average, best, and worst performances (with shading in between) across the 50 replicates every 2000 generations.}
\label{fig:res:mvc-exploitation}
\end{figure}

Here, we discuss the basic valley-crossing diagnostic (diagnostic 5) in detail, without the additional complications associated with constrained exploitation or parallel exploration.
We report all results for the full set of valley-crossing diagnostics in our supplemental material \citep{supplemental_material}.
Integrating fitness valleys with either the ordered exploitation or multi-path exploration diagnostics generally led to stagnating populations and did not produce any surprising results.


The basic valley-crossing diagnostic employs a search space with independent gradients that possess identical valley structures (Figure \ref{fig:sawtooth-transformation}b).
We found that all selection schemes, except lexicase, outperformed the random control when comparing the best performance earned throughout an evolutionary run (Figure \ref{fig:res:mvc-exploitation}b; Wilcoxon rank-sum tests: $p<10^{-4}$).
Of the eight selection schemes, the two fitness sharing variations managed to cross the greatest number of (and widest) fitness valleys and significantly outperformed all other schemes (\cite{supplemental_material}; Wilcoxon rank-sum tests: $p<10^{-4}$). 
We did not detect significant performance differences between fitness sharing variants (Wilcoxon rank-sum test: $p>0.05$). 
Because fitness sharing penalizes a solution's fitness based on its similarity to the rest of the population, solutions crowded together on a particular fitness peak will often have similar or worse fitness to a solution heading through a nearby fitness valley.
Indeed, this shared fitness mechanism was originally designed to help promote valley crossing \citep{goldberg1987genetic}.




Lexicase selection excels at simultaneously exploring independent gradients without completely sacrificing its ability to exploit.
However, it possesses no mechanism to steer solutions through fitness valleys.
While lexicase shuffles a set of test cases to generate solutions that specialize on different combinations of test cases, a version of elitism is still maintained at the level of individual test cases.
Indeed, lexicase's inability to cross valleys is illustrated in these results, as it performed worse than all other selection schemes  (Figure \ref{fig:res:mvc-exploitation}b; Wilcoxon rank-sum tests: $p<10^{-4}$), and no differences were detected when compared to our random control (Wilcoxon rank-sum tests: $p>0.05$).
These results demonstrate a limitation of lexicase, where it performed well on all of the diagnostics without fitness valleys.
While no schemes managed to reach high-performing solutions, schemes less reliant on elitism (fitness sharing, nondominated sorting, and novelty search) show more potential to continue to improve. 

%% file: Text/conclusion.tex
\section{Conclusion}

\begin{figure}[h]
\centering
\includegraphics[width=\textwidth]{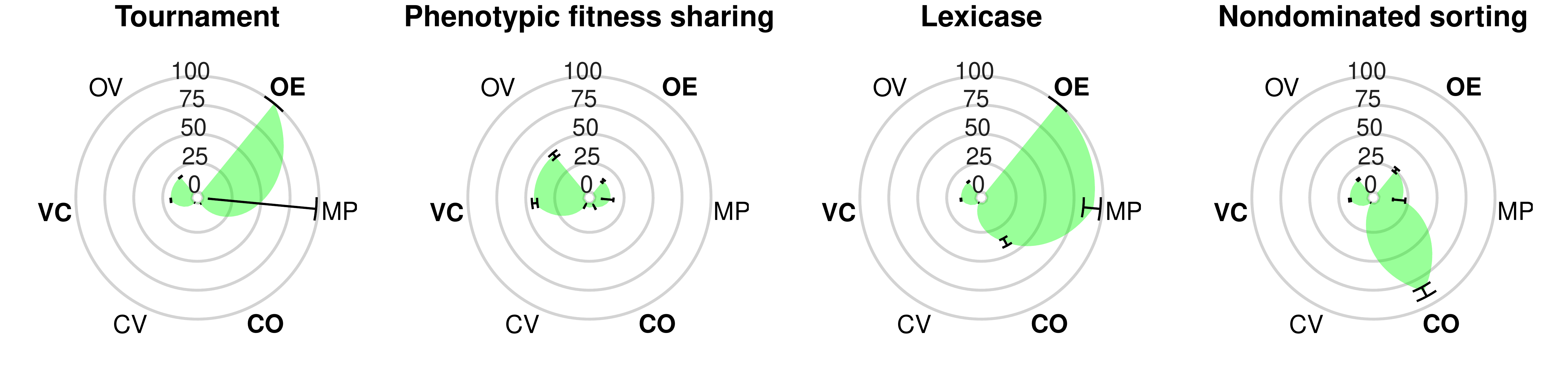}
\caption{
\textbf{Diagnostic profiles for tournament selection, phenotypic fitness sharing, lexicase, and nondominated sorting.} 
We report the best performance found throughout an evolutionary search for the exploitation rate with valleys (\textbf{VC}), ordered exploitation with (OV) and without (\textbf{OE}) valleys, and multi-path exploration (MP) diagnostics, and the satisfactory trait coverage found in the final population for the contradictory objectives diagnostic (\textbf{CO}) and its valley crossing variant (CV).
We plot the average across the 50 replicates, with error bars between the best and worst data.
}
\label{fig:con:radar-profiles}
\end{figure}

Our handcrafted diagnostics allow us to focus on specific problem characteristics, providing direct intuition about how selection schemes handle different types of search spaces.
Ideally, before a new scheme is introduced, it should generate promising results on some set of diagnostics, as compared to existing schemes. 
As an example for how to make this comparison, Figure \ref{fig:con:radar-profiles} illustrates a ``diagnostic profile'' for four selection schemes, highlighting major differences among them.
The current version of this diagnostic suite is merely a starting point that we envision growing as we and others in the evolutionary computing community identify additional salient characteristics of search spaces to incorporate. 
We expect that new diagnostics will often be inspired by selection schemes that demonstrate successes on real-world problems that are not explained by existing diagnostics.


Our diagnostics do face some similar issues to benchmark suites, as discussed in Section \ref{sec:intro}.
Specifically, diagnostics must still be configured (e.g., dimensionality, mutation rate, population size, etc.).
More research is needed to understand how sensitive selection schemes are to diagnostic configurations, but we expect the general results to be robust. 
While it may be possible to configure a diagnostic to be biased toward a particular selection scheme, a consistent use of the same configuration across different diagnostics and different schemes should minimize this issue. 
In terms of statistical significance, too few replicates can still cause problems, but our diagnostics are fast to evaluate so replicates can be quickly generated. 

Overall, our results are consistent with previous work.
Truncation and tournament selection were heavily exploitative with poor capacities for either parallel or valley-crossing exploration.
Novelty search was purely exploratory with no mechanism for exploitation.
Nondominated sorting excelled at managing multiple contradictory objectives, but did not exploit gradients or cross valleys well. 
Lexicase selection effectively balanced exploitation with parallel exploration, performing reasonably well across all diagnostics that did not incorporate fitness valleys.
The difficulty of each diagnostic increased with the addition of fitness valleys, where fitness sharing was the only selection scheme to make significant progress when valleys were added.
Of particular note, lexicase selection performed worse than random selection when contending with fitness valleys, which is surprising given lexicase's pedigree of problem-solving success across a wide range of domains.  
These results indicate the need for more sophisticated schemes to contend with challenging combinations of search-space characteristics.

Because the results for each diagnostic are heavily dependent on selection scheme configurations, we included additional replicates that explore different parameter configurations in our supplemental material \citep{supplemental_material}.
Overall, our results emphasize the importance of choosing the appropriate scheme for a given problem, as each of the schemes investigated here exhibited distinct trade-offs between different problem characteristics.
These diagnostics also help identify missing combinations of capabilities, providing a clear direction for designing new selection schemes.


One approach to improve success on the more challenging diagnostics is to use more than one selection scheme to identify parents so that a mix of their exploitation and exploration abilities can be generated. 
For example, truncation selection can be used for exploitation, and novelty search can be used for exploration.
Another approach is to combine techniques used by different selection schemes.
For example, lexicase selection performs well on all diagnostics that do not incorporate fitness valleys, while fitness sharing contends best with diagnostics that do incorporate valleys.
It may be possible to overcome the limitations that each scheme possesses by combining fitness sharing and lexicase selection.
Ultimately, our diagnostics help us validate these approaches, test different variations of selection methods, and investigate schemes with more control than previously possible with standard benchmarking approaches.